\documentclass[authoryear,preprint,12pt]{elsarticle}
\usepackage{amsmath,amssymb,amsthm,amsfonts,fourier,booktabs,algorithm,algorithmic,blkarray,xparse,soul,comment,hyperref,nicefrac,xcolor,bm,tikz,pgfplots,subfigure,multirow,arydshln} 
\newtheorem{example}{Example}
\newtheorem{theorem}{Theorem}

\newtheorem{proposition}{Proposition}

\pgfplotsset{compat=1.17} 
\usepgfplotslibrary{statistics}
\usetikzlibrary{pgfplots.statistics,arrows,plotmarks,decorations.markings,trees,shapes,pgfplots.groupplots,automata,circuits}
\tikzset{dot/.style = {circle, fill, minimum size=#1,inner sep=0pt, outer sep=0pt, fill, circle},dot/.default = 6pt}
\tikzset{dot2/.style = {circle, fill, color=black!40,minimum size=6pt,inner sep=0pt, outer sep=0pt, fill, circle}}
\tikzstyle{a}=[->,>=stealth,dashed]
\tikzstyle{a2}=[->,>=stealth]
\tikzstyle{nodo}=[ellipse,draw=black!100,fill=black!0,line width=.7pt,minimum width=1.2cm,minimum height=0.8cm,text width=1.2cm,text centered]
\tikzstyle{nodo2}=[ellipse,draw=black!100,fill=black!10,line width=.7pt,minimum width=1.2cm,minimum height=0.8cm,text width=1.2cm,text centered]
\tikzstyle{nodo3}=[ellipse,draw=black!100,fill=black!30,line width=.7pt,minimum width=1.2cm,minimum height=0.8cm,text width=1.2cm,text centered]
\tikzstyle{arco}=[draw=black!80,line width=.7pt, postaction={decorate}, decoration={markings,mark=at position 1.0 with {\arrow[ draw=black!80,line width=.7pt]{>}}}]
\journal{International Journal of Approximate Reasoning}
\begin{document}
\begin{frontmatter}
\title{Learning to Bound Counterfactual Inference from Observational, Biased and Randomised Data}
\author[IDSIA]{Marco Zaffalon}
\ead{marco.zaffalon@idsia.ch}
\author[IDSIA]{Alessandro Antonucci\corref{cor1}}
\ead{alessandro.antonucci@idsia.ch}
\cortext[cor1]{Corresponding author}
\author[UAL]{Rafael Cabañas}
\ead{rcabnas@ual.es}
\author[IDSIA]{David Huber}
\ead{david.huber@idsia.ch}
\address[IDSIA]{IDSIA, Lugano (Switzerland)}
\address[UAL]{Department of Mathematics, University of Almería, Almería (Spain)}
\tnoteref{label1}
\begin{abstract}
We address the problem of integrating data from multiple, possibly biased, observational and interventional studies, to eventually compute counterfactuals in structural causal models. We start from the case of a single observational dataset affected by a selection bias. We show that the likelihood of the available data has no local maxima. This enables us to use the causal expectation-maximisation scheme to compute approximate bounds for partially identifiable counterfactual queries, which are the focus of this paper. We then show how the same approach can solve the general case of  multiple datasets, no matter whether interventional or observational, biased or unbiased, by remapping it into the former one via graphical transformations. Systematic numerical experiments and a case study on palliative care show the effectiveness and accuracy of our approach, while hinting at the benefits of integrating heterogeneous data to get informative bounds in case of partial identifiability.
\end{abstract}
\end{frontmatter}
\section{Introduction}
Consider the data in Table~\ref{tab:study}. They represent the results of an artificial case study from \citet{mueller2022}, where patients affected by a deadly disease test a drug that might help them. The table reports two independent studies carried out on different groups of patients: a randomised control trial (first eight rows) and an observational study (last eight rows). From those data, \citet{mueller2022} compute a typical counterfactual such as the \emph{probability of necessity and sufficiency} (PNS), which is the proportion of people that would recover if treated and die otherwise. They can do this analytically by conditioning on gender. For males they eventually obtain that $\mathrm{PNS}\in[0.28,
0.49]$ if only the randomised trial is considered, whereas the sharp value $\mathrm{PNS}=0.49$ is obtained when observational data are also taken into account. Analogous results are obtained for females.

\begin{table}[htp!]
\centering
\begin{tabular}{llllr}
\toprule
Study&Treatment&Gender&Survival&Counts\\
\midrule
\multirow{8}{*}{interventional}&do(drug)&female&survived&489\\
&do(drug)&female&dead&511\\
&do(drug)&male&survived&490\\
&do(drug)&male&dead&510\\
&do(no drug)&female&survived&210\\
&do(no drug)&female&dead&790\\
&do(no drug)&male&survived&210\\
&do(no drug)&male&dead&790\\
\hline
\multirow{8}{*}{observational}&drug&female&survived&378\\
&drug&female&dead&1022\\
&{\color{black!30}{drug}}&{\color{black!30}{male}}&{\color{black!30}{survived}}&{\color{black!30}{980}}\\
&{\color{black!30}{drug}}&{\color{black!30}{male}}&{\color{black!30}{dead}}&{\color{black!30}{420}}\\
&{\color{black!30}{no drug}}&{\color{black!30}{female}}&{\color{black!30}{survived}}&{\color{black!30}{420}}\\
&{\color{black!30}{no drug}}&{\color{black!30}{female}}&{\color{black!30}{dead}}&{\color{black!30}{180}}\\
&no drug&male&survived&420\\
&no drug&male&dead&180\\
\bottomrule
\end{tabular}
\caption{Data from interventional and observational studies on the potential effects of a drug on patients affected by a deadly disease.}\label{tab:study}
\end{table}

This example illustrates the main traits of the problem we are going to address in this paper: first and foremost that the computation of counterfactuals is often only \emph{partially identifiable}, in the sense that we can at best obtain probabilistic bounds for them; but also that joining observational and randomised studies (under the guidance of a so-called \emph{structural causal model}) can strengthen the results, i.e., narrow the bounds. 

What the original example instead does not take into account is the possible presence of a \emph{selection bias}, which is the systematic exclusion of a certain subpopulation from the sample. To account for this, we assume here that half of the data (the ones about treated males and untreated females, denoted in grey in the table) from the observational study is missing for a problem with the communication protocol. The challenge is to produce correct counterfactual bounds for the overall population given the partial view induced by the biased data (in what follows, we call these data under selection bias just `biased' for brevity). 

In this paper we shall give general methods to solve problems as above: that is, the computation of counterfactuals with structural causal models where available data are a mix of observational and interventional (e.g., based on randomised studies) datasets, possibly subject to selection bias. We refer to these data as  \emph{heterogeneous}. 

Such a setting essentially coincides with the general problem of information fusion as defined by \citet{bareinboim2016causal}. This, and nearly all other works in the literature, have however focused on the special case of identifiable queries, namely, those that can be reduced to probabilistic expressions. The only apparent exception is the recent work by \cite{zhang2021} that showcases the application of MCMC techniques to solve a problem of information fusion without selection bias. Selection bias in itself has been studied in the causal literature for long (e.g., \citet{cooper1995causal,pearl2012solution,bareinboim2012controlling,bareinboim2015recovering}) but the unidentifiable case is still basically unexplored.

In contrast with the trend in the literature, this paper will entirely focus on the general case of partial identifiability. In particular, after reviewing the necessary background material about structural causal models and Bayesian networks in  Section~\ref{sec:background}, we shall quickly review our EM-based scheme (we call it EMCC) for observational datasets in Section~\ref{sec:emcc}. The EMCC is an algorithm that takes a structural causal model in input, along with observational data, and allows us to approximately compute bounds for any counterfactual. The EMCC will be extended to the case of biased datasets in Section~\ref{sec:s-emcc} and to multiple studies is Section~\ref{sec:multi-db}. We shall present in Section~\ref{sec:experiments} a numerical validation on a synthetic benchmark and a case study on palliative care involving a biased observational dataset. Conclusions and outlooks will be reported in Section~\ref{sec:conclusions}.

\section{Background on Bayesian Networks and Structural Causal Models}\label{sec:background}
A generic variable $X$ is assumed to take values from a finite set $\Omega_X$. A probability mass function (PMF) over $X$ is denoted as $P(X)$. Given variables $Y$ and $X$, a conditional probability table (CPT) $P(Y|X)$ is a collection of PMFs over $Y$ indexed by the values of $X$, i.e., $\{P(Y|X)\}_{x\in\Omega_X}$. If all PMFs in a CPT are \emph{degenerate}, i.e., $\{0,1\}$-valued, we say that also the CPT is degenerate. 

A \emph{structural equation} (SE) $f$ associated with variable $Y$ and based on the input variable(s) $X$, is a surjective function $f:\Omega_{X} \to \Omega_Y$ that determines the value of $Y$ from that of $X$. Such an SE induces the degenerate CPT $P(Y|X)$ via $P(y|x):=\llbracket f(x)=y \rrbracket$ for each $x\in\Omega_X$ and $y\in\Omega_Y$, where $\llbracket \cdot \rrbracket$ denotes the Iverson brackets that take value one if the statement inside the brackets is true and zero otherwise.

Consider a joint variable $\bm{X}:=(X_1,\ldots,X_n)$ and a directed acyclic graph $\mathcal{G}$ whose nodes are in a one-to-one correspondence with the variables in $\bm{X}$ (we use a node in $\mathcal{G}$ and its corresponding variable interchangeably). Given $\mathcal{G}$, a Bayesian network (BN) is a collection of CPTs  $\{P(X_i|\mathrm{Pa}_{X_i})\}_{i=1}^n$, where $\mathrm{Pa}_{X_i}$ denotes the \emph{parents} of $X_i$, i.e., the direct predecessors of $X_i$ according to $\mathcal{G}$. A BN induces a joint PMF $P(\bm{X})$ that factorises as $P(\bm{x})=\prod_{i=1}^n P(x_i|\mathrm{pa}_{X_i})$, for each $\bm{x}\in\Omega_{\bm{X}}$, where $(x_i,\mathrm{pa}_{X_i})\sim \bm{x}$, i.e., $x_i$ and $\mathrm{pa}_{X_i}$ are the values of $X_i$ and $\mathrm{Pa}_{X_i}$ consistent with $\bm{x}$ for each $i=1,\ldots,n$.

Now consider two joint variables $\bm{U}$ and $\bm{V}$, which we respectively refer to as \emph{exogenous} and \emph{endogenous}. A collection of SEs $\{f_V\}_{V\in\bm{V}}$ such that, for each $V\in\bm{V}$ the input variables of $f_V$ are in $(\bm{U},\bm{V})$, is called a \emph{partially specified} structural causal model (PSCM). A PSCM $M$ induces the specification of a directed graph $\mathcal{G}$ with nodes in a one-to-one correspondence with the variables in $(\bm{U},\bm{V})$ and such that there is an arc between two variables if and only if the first variable is an input variable for the SE of the second. The exogenous variables are therefore root nodes of $\mathcal{G}$. We focus on \emph{semi-Markovian} PSCMs, i.e., those PSCMs that lead to acyclic graphs. 

In a PSCM $M$ we obtain a joint state of $\bm{V}$ from a (joint) state of $\bm{U}$ by applying the SEs of $M$ consistently with a topological order for $\mathcal{G}$. A \emph{fully specified} structural causal model (FSCM) is just a PSCM $M$ paired with a collection of marginal PMFs, one for each exogenous variable. As SEs induce (degenerate) CPTs, an FSCM defines a BN based on $\mathcal{G}$ whose joint PMF factorises as:
\begin{equation}\label{eq:fulljoint}
P(\bm{u},\bm{v})= \prod_{U\in\bm{U}} \theta_{u} \cdot \prod_{V\in\bm{V}} P(v|\mathrm{pa}_V)\,,
\end{equation}
for each $\bm{u}\in\Omega_{\bm{U}}$ and $\bm{v}\in\Omega_{\bm{V}}$, $(u,v,\mathrm{pa}_V)\sim(\bm{u},\bm{v})$ and where $\mathrm{Pa}_V$ are the parents of $V$ according to $\mathcal{G}$ (i.e., the inputs of SE $f_V$), while $\theta_u$ denotes the true but unknown chances for $U=u$, to be considered for each $u\in\Omega_U$ and $U\in\bm{U}$. Notation $\theta_U:=(\theta_u)_{u\in\Omega_U} \in \Delta_U$ is similarly used for the whole PMF, with $\Delta_U$ denoting the corresponding probability simplex. We extend this notation to the joint variable $\bm{U}$ via $\theta_{\bm{U}}:=\times_{U\in\bm{U}} \theta_U\in\times_{U\in\bm{U}} \Delta_U=:\Delta_{\bm{U}}$. We shall instead use $P(u)$ for an estimate of $\theta_u$ and likewise for PMFs.

Given the graph $\mathcal{G}$ of a PSCM (or FSCM) $M$, obtain $\mathcal{G}'$ by removing from $\mathcal{G}$ any arc connecting pairs of endogenous variables. Let $\{\mathcal{G}_c\}_{c\in\mathcal{C}}$ denote the connected components of $\mathcal{G}'$. The \emph{c-components} of $M$ are the elements of the partition $\{\bm{V}^{(c)}\}_{c\in\mathcal{C}}$ of $\bm{V}$, where $\bm{V}^{(c)}$ denotes the endogenous nodes in $\mathcal{G}_c$, for each $c\in\mathcal{C}$ \citep{tian2002studies}. This procedure also induces a partition of $\bm{U}$, similarly denoted as $\{\bm{U}^{(c)}\}_{c\in\mathcal{C}}$. Moreover, for each $c\in\mathcal{C}$, let $\bm{W}^{(c)}$ denote the union of the endogenous parents of the nodes in $\bm{V}^{(c)}$ and $\bm{V}^{(c)}$ itself. Finally, for each $V\in \bm{V}^{(c)}$, obtain $\bm{W}_V$ by removing from $\bm{W}^{(c)}$ the nodes topologically following $V$ and $V$ itself (note that in the notation we dropped the index $c$ as this can be implicitly retrieved from $V$). \citet{tian2002studies} shows that the joint PMF $P(\bm{V})$ obtained by marginalising the exogenous variables from the joint PMF in Equation~\eqref{eq:fulljoint} is a BN, to be called here \emph{endogenous}, that factorises as follows:
\begin{equation}\label{eq:empirical}
P(\bm{v})=\prod_{V\in\bm{V}} P(v|\bm{w}_V)\,,
\end{equation}
for each $\bm{v}\in\Omega_{\bm{V}}$, with $(v,\bm{w}_V)\sim \bm{v}$ for each $V\in\bm{V}$. 

In FSCMs, the CPTs in the right-hand side of Equation~\eqref{eq:empirical} can be computed through standard BN inference algorithms by simply regarding the FSCM as a BN. With PSCMs, assuming the availability of a dataset $\mathcal{D}$ of endogenous observations, we might also define an endogenous BN with the same factorisation and whose CPTs are directly assessed from $\mathcal{D}$.

Observational queries in FSCMs can be addressed in the endogenous BN, and the same can be done for PSCMs assuming the availability of the dataset $\mathcal{D}$ of endogenous observations. To perform causal inference, \emph{interventions} denoted as $\mathrm{do}(\cdot)$ should be considered instead. In an FSCM or PSCM $M$, given $V \in \bm{V}$ and $v\in\Omega_V$, $\mathrm{do}(V=v)$ simulates a physical action on $M$ forcing $V$ to take the value $v$. The original SE $f_V$ should be consequently replaced by a constant map $V=v$. Notation $M_v$ is used for such a modified model, whose graph is obtained by removing from $\mathcal{G}$ the arcs entering $V$, and for which evidence $V=v$ is considered. In an FSCM $M$, given $V,W\in\bm{V}$ and $v\in\Omega_V$, $P(w|\mathrm{do}(v))$ denotes the conditional probability of $W=w$ in the post-intervention model, i.e., $P'(w|v)$, where $P'$ is the joint PMF induced by $M_v$. 

A more general setup is provided by \emph{counterfactual} queries, where the same variable may be observed as well as subject to intervention, albeit in distinct \emph{worlds}. In mathematical parlance, if $\bm{W}$ are the queried variables, $\bm{V}'$ the observed ones and $\bm{V}''$ the intervened ones, we write the query by $P(\bm{W}_{\bm{v}''}|\bm{v}')$ with possibly $\bm{V}'\cap \bm{V}''\neq \emptyset$. Popular examples of counterfactual queries involving two endogenous Boolean variables $X$ and $Y$ of an FSCM are the \emph{probability of necessity} (PN), i.e., the probability that event $Y$ would not have occurred by disabling $X$, given that $X$ and $Y$ did in fact occur. This corresponds to $P(Y_{X=0}=0|X=1,Y=1)$. Similarly the probability of \emph{sufficiency} (PS) $P(Y_{X=1}=1|X=0,Y=0)$ and \emph{necessity and sufficiency} (PNS) $P(Y_{X=1}=1,Y_{X=0}=0)$ are often considered.

Computing counterfactual queries in an FSCM may be achieved via an auxiliary structure called a \emph{twin network}  \citep{balke1994counterfactual}. This is simply an FSCM where the original endogenous nodes (and their SEs) have been duplicated, while remaining affected by the same exogenous variables. More general and compact structures (e.g., involving more than two copies of the same endogenous node) can be also considered \citep{shpitser2007counterfactuals}. Computing a counterfactual in the twin network of an FSCM is analogous to what is done with interventional queries provided that interventions and observations are associated with distinct copies of the same variable. BN inference eventually allows one to compute the counterfactual query in such an augmented model. 

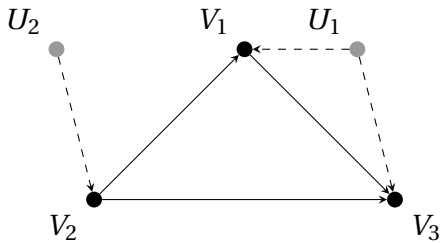
\begin{figure}[htp!]
\centering
\begin{tikzpicture}[scale=1.0]
\node[dot2,label=above left:{$U_1$}] (u1)  at (2.5,1) {};
\node[dot2,label=above left:{$U_2$}] (u2)  at (-1.5,1) {};
\node[dot,label=above left:{$V_1$}] (v1)  at (1,1) {};
\node[dot,label=below left:{$V_2$}] (v2)  at (-1,-1) {};
\node[dot,label=below right:{$V_3$}] (v3)  at (3.0,-1) {};
\draw[a] (u2) -- (v2);
\draw[a] (u1) -- (v1);
\draw[a] (u1) -- (v3);
\draw[a2] (v2) -- (v1);
\draw[a2] (v1) -- (v3);
\draw[a2] (v2) -- (v3);
\end{tikzpicture}
\caption{The graph of a structural causal model over three endogenous variables (black nodes) with two exogenous variables (grey nodes). The model is used to describe the data in Table~\ref{tab:study} with $V_1$ corresponding to \emph{Treatment}, $V_2$ to \emph{Gender} and $V_3$ to \emph{Survival}.
The exogenous variable $U_1$ acts as a \emph{confounder} for $V_1$ and $V_3$, this being sufficient to have results consistent with those of \citet{mueller2022}, which are obtained by conditioning on $V_2$.}\label{fig:scm}
\end{figure}

\section{Partial Identifiability (with a Single Unbiased Observational Dataset)}\label{sec:emcc}
In this section we formalise the fundamental notion of partially identifiable query and review the basics of the EM-based scheme proposed by \citet{zaffalon2021} to address queries of this kind. Such an EM procedure considers the case of a single, purely observational and unbiased, dataset. An extension to the case of (single and observational) biased datasets is in Section~\ref{sec:s-emcc}, while the fully general case of multiple, observational and interventional, unbiased or biased, datasets is addressed in Section~\ref{sec:multi-db}.

As already noticed in the previous section, the general problem of computing counterfactuals can be reduced to standard BN inference in FSCMs. Yet exogenous variables are typically latent; their marginal PMFs being unavailable leaves us with PSCMs. 

Say that we have a PSCM $M$ over variables $(\bm{U},\bm{V})$ together with a dataset $\mathcal{D}$ of endogenous observations. This permits quantifying the endogenous BN from which any observational query can be computed. The do-calculus \citep{pearl2009causality} instead permits reducing \emph{identifiable} interventional queries on PSCMs to observational ones, which are possibly then computed in the endogenous BN. The remaining queries are called here \emph{partially} identifiable. 

To start approaching partially identifiable queries in the PSCM $M$, from $\mathcal{D}$ we first compute the endogenous BN inducing the endogenous joint PMF $P(\bm{V})$. To be consistent with the endogenous PMF, the exogenous PMF $\theta_{\bm{U}}$ of an FSCM should satisfy:
\begin{equation}\label{eq:ident}
\sum_{\bm{u}\in\Omega_{\bm{U}}} 
\left[ 
\prod_{U\in\bm{U}} \theta_u 
\cdot \prod_{V\in\bm{V}} P_M(v|\mathrm{pa}_V) 
\right] = P(\bm{v})\,,
\end{equation}
for each $\bm{v}\in\Omega_{\bm{V}}$, with $(u,v,\mathrm{pa}_V)\sim (\bm{u},\bm{v})$ and $\theta_u \sim \theta_{\bm{U}}$. If such a $\theta_{\bm{U}}$ exists, we say that the endogenous BN is \emph{$M$-compatible} with the original PSCM \citep{zaffalon2021}. In general there are multiple FSCMs satisfying Equation~\eqref{eq:ident}, that is, there is a set $\Theta_{\bm{U}}\subseteq\Delta_{\bm{U}}$ of elements $\theta_{\bm{U}}$ that satisfies Equation~\eqref{eq:ident}. To compute a partially identifiable causal query in PSCMs, we should therefore consider {all} $\theta_{\bm{U}}\in\Theta_{\bm{U}}$ and compute the query for each of them. The result of the partially identifiable query is understood as the interval spanned by the outcomes obtained on all the ($M$-)compatible FSCMs.

To practically pursue that, it is useful to consider the log-likelihood of $\mathcal{D}$ from a generic FSCM based on $M$, i.e.,
\begin{equation}\label{eq:liku}
LL(\theta_{\bm{U}}) := \sum_{\bm{v}\in\mathcal{D}} \log \sum_{\bm{u}\in\Omega_{\bm{U}}} 
\left[ \prod_{U\in\bm{U}} {\theta}_{{u}} \cdot \prod_{V\in\bm{V}} P(v|\mathrm{pa}(V)) \right],\,
\end{equation}
with $(u,v,\mathrm{pa}(V)) \sim (\bm{u},\bm{v})$ and ${\theta}_{{u}} \sim \theta_{\bm{u}}$. Interestingly, this function has a single global maximum and achieving such a maximum is an equivalent condition to identify the  ($M$-)compatible FSCMs, as stated by the following result.

\begin{proposition}[\citeauthor{zaffalon2021}, \citeyear{zaffalon2021}]\label{prop:iff}
The log-likelihood in Equation~\eqref{eq:liku} has no local maxima and achieving its global maximum at $\theta_{\bm{U}}$ is an equivalent condition for $\theta_{\bm{U}}\in\Theta_{\bm{U}}$.
\end{proposition}

The lack of local maxima allows us to use an EM scheme, iterated with different random initialisations, to sample points from $\Theta_{\bm{U}}$. Given a PSCM $M$ and a dataset of (unbiased) endogenous observations $\mathcal{D}$, Algorithm~\ref{alg:cem} yields an FSCM compatible with the PSCM (provided that there is one): such an FSCM is obtained by adding PMFs to the exogenous variables of the PSCM in such a way that the resulting model can generate the distribution of the available data.

In practice, given an initialisation $\{P_0(U)\}_{U\in\bm{U}}$, the EM algorithm consists in regarding the posterior probability $P_0(u|\bm{v})$ as a pseudo-count for $(u,\bm{v})$, for each $\bm{v}\in\mathcal{D}$,  $u\in\Omega_U$ and $U\in\bm{U}$ (E-step). A new estimate $P_1(u):=\sum_{\bm{v}\in\mathcal{D}} \frac{P_0(u|\bm{v})}{|\mathcal{D}|}$ is consequently obtained (M-step). The scheme is iterated until convergence. Subroutine ${\tt initialisation}$ provides a random initialisation of the exogenous PMFs (line 1). The iterative procedure stops when the log-likelihood achieves a maximum (line 3).

Note that the numerical value of the global maximum of the log-likelihood can be proved to coincide with the log-likelihood of the dataset $\mathcal{D}$ from the endogenous BN whose parameters are quantified with the frequencies in $\mathcal{D}$ itself \citep{zaffalon2021}. Such a value can be used to detect the two special cases in which the EMCC does not converge to the global maximum. The first is the convergence to a saddle point: this is an unstable point and a global maximum replacing that value can be easily obtained by a small perturbation in the initialisation. On the other side, if there is no convergence to the global maximum irrespectively of the initialisation, which means that the endogenous BN is not $M$-compatible. As discussed by \citet{zaffalon2021} this is due to a wrong modelling or to insufficient data and makes inferences unreliable irrespectively of the procedure adopted to compute them.

Computing a causal query in the FSCM returned by a single EMCC run gives the exact value of the query in the identifiable case, which then corresponds to a point of the interval for the partially identifiable case represented by the PSCM. An inner approximation of such an interval is eventually achieved by iterating the EM scheme above with variable initialisation values. This corresponds to sampling the space of FSCMs compatible with the given PSCM.

\begin{algorithm}[htp!]
\caption{EMCC: given SCM $M$ and dataset $\mathcal{D}$ returns $\{P(U)\}_{U\in\bm{U}}$.}
\begin{algorithmic}[1]
\STATE $\{P_0(U)\}_{U\in\bm{U}} \leftarrow {\tt initialisation}(M)$
\STATE $t \leftarrow 0$
\WHILE{$LL(\{P_{t+1}(U)\}_{U\in\bm{U}})>LL(\{P_t(U)\}_{U\in\bm{U}})$}
\FOR{$U\in \bm{U}$}
\STATE $P_{t+1}(U) \leftarrow |\mathcal{D}|^{-1}\sum_{\bm{v} \in \mathcal{D}} P_t(U|\bm{v})$
\STATE $t \leftarrow t+1$
\ENDFOR
\ENDWHILE
\end{algorithmic}\label{alg:cem}
\end{algorithm}

A demonstrative example of our EM procedure is reported here below. Here and in the other examples we consider \emph{canonical} SEs \citep{zhang2021}, this meaning that the cardinality of the exogenous variables allows to have them enumerating all the possible deterministic relations between the endogenous output and its endogenous inputs.

\begin{example}\label{ex:unbiased}
Consider the observational study associated with the last eight rows in Table~\ref{tab:study} and a PSCM $M$ based on the causal graph in Figure~\ref{fig:scm} with a canonical specification of the SEs (this requires $|\Omega_{U_1}|=64$ and $|\Omega_{U_2}|=2$). We are interested in a counterfactual query corresponding to the PNS for $V_1$ (i.e., \emph{Treatment}) as subject of the intervention and $V_3$ (i.e., \emph{Survival}) as queried variable. Each run of Algorithm~\ref{alg:cem} corresponds to an FSCM. The interval spanned by the corresponding PNSs after $r=300$ runs and a rounding to the second digit is $[0.00,0.43]$. In the other examples discussed in the rest of the paper we consider, with different setups, the same counterfactual query computed with the same number of runs and rounding strategy. The script to reproduce all these simulations is available together with the material released to reproduce the experiments (see Section~\ref{sec:experiments}).
\end{example}

Other approaches have recently been proposed to approximate partially identifiable bounds  \citep{zaffalon2020,duarte2021,zhang2021}. Yet, these appears are focused on the case of unbiased data, and no viable technique to compute general counterfactual inference under selection bias seems to be available. This is the focus of the next section, which shows how the EMCC of \citet{zaffalon2021} has a natural extension to problems of selection bias.

\section{Partial Identifiability under Selection Bias}\label{sec:s-emcc}
To model the effect of selection bias, we define a \emph{selector} $S$, i.e., a Boolean variable that is true for selected states of $\bm{V}$ and false otherwise. We consider deterministic selectors, i.e., $S:=g(\bm{V})$ with $g:\Omega_{\bm{V}}\to \{0,1\}$. E.g., for the observational data in Table~\ref{tab:study}, if we regard \emph{drug} and \emph{female} as the first states of the variables \emph{Gender} and \emph{Treatment}, the selector is true if and only if the two variables are in the same state. $S$ can be embedded in a PSCM $M$ as a common child of the endogenous variables, with $g$ being the SE of $S$, thus acting as an additional endogenous variable with endogenous parents only. Notation $M^S$ is used to denote such an augmented PSCM obtained from $M$.

Selector $S$ induces a partition among the records of $\mathcal{D}$. The endogenous values are assumed to become missing on the records such that $S=0$, while remaining available if $S=1$. Let denote the corresponding datasets as $\mathcal{D}_{S=0}$ and $\mathcal{D}_{S=1}$. In terms of cardinalities, if $N_{S=0}:=|\mathcal{D}_{S=0}|$ and $N_{S=1}:=|\mathcal{D}_{S=1}|$, then $N_{S=0}+N_{S=1}=N=:|\mathcal{D}|$. Note that $\mathcal{D}_{S=0}$ contains $N_{S=0}$ identical records. Figure~\ref{fig:sb} depicts an example of the partition for the biased observational data of Table~\ref{tab:study}.

\begin{figure}[htp!]
\centering
\begin{tikzpicture}
\node (tab) at (4,1) {
\begin{tabular}{ccccc}
\toprule
$V_1$&$V_2$&$V_3$&$S$&\#\\
\midrule
0&0&0&0&378\\
0&0&1&0&1022\\
0&1&0&1&980\\
0&1&1&1&420\\
1&0&0&1&420\\
1&0&1&1&180\\
1&1&0&0&420\\
1&1&1&0&180\\
\bottomrule
\end{tabular}};
\node (tab1) at (11,2.1) {
\begin{tabular}{cc ccc cc}
\hline 
$U_1$&$U_2$&$V_1$&$V_2$&$V_3$&$S$&\#\\
\hline 
*&*& 0&1&0&1&378\\
*&*& 0&1&1&1&1022\\
*&*& 1&0&0&1&420\\
*&*& 1&0&1&1&180\\
\hline 
\end{tabular}};
\node (tab0) at (11,-1) {
\begin{tabular}{ccccccc}
\hline 
$U_1$&$U_2$&$V_1$&$V_2$&$V_3$&$S$&\#\\
\hline 
*&*& *&*&*& 0&1000\\
\hline 
\end{tabular}};
\draw[a2] (tab) -- (tab1);
\draw[a2] (tab) -- (tab0);
\end{tikzpicture}
\caption{Selecting the data of the observational study in Table~\ref{tab:study}.}\label{fig:sb}
\end{figure}

To address partially identifiable queries under selection bias, let us first reformulate the notion of $M$-compatibility for $M^S$. On the selected states the compatibility condition is like in the unbiased case, and we consequently require Equation~\eqref{eq:ident} to be satisfied for each $\bm{v}$ such that $g(\bm{v})=1$, where the joint endogenous probability on the right-hand side is computed by an endogenous BN whose CPTs are obtained from the frequencies in $\mathcal{D}_{S=1}$. For the unselected states, as the only endogenous observation is $S=0$, we require instead:
\begin{equation}\label{eq:sm_compat11}
\sum_{\bm{v} : g(\bm{v})=0}
\sum_{\bm{u} \in \Omega_{\bm{U}}} 
\left[ \prod_{U\in\bm{U}} \theta_u \cdot \prod_{V\in\bm{V}}
\llbracket f_V(\mathrm{pa}_V)-v \rrbracket 
\right] = P(S=0)\,,
\end{equation}
where, like for the endogenous joint probabilities, also the marginal probabilities of the selector are obtained from the (cardinalities of the) data, e.g., $P(S=0) \simeq N_{S=0}/N$.

The task of detecting the set $\Theta_{\bm{U}}$ of elements $\theta_{\bm{U}}$ satisfying the above $M$-compatibility conditions can be reduced to a log-likelihood maximisation as in the unbiased case. The log-likelihood of $\mathcal{D}_S:=\mathcal{D}_{S=0} \cup \mathcal{D}_{S=1}$ from an FSCM based on $M^S$  can be written as:
\begin{equation}\label{eq:llu}
LL(\theta_{\bm{U}}):=
LL_{S=0}(\theta_{\bm{U}})+
LL_{S=1}(\theta_{\bm{U}})\,,    
\end{equation}
with:
\begin{eqnarray}\label{eq:ll0}
LL_{S=0}(\theta_{\bm{U}}) &:=&
N_{S=0} \log \sum_{\bm{v}:g(\bm{v})=0} \sum_{\bm{u} \in \Omega_{\bm{U}}}
\left[\prod_{U\in\bm{U}} \theta_{u} \cdot 
\prod_{V\in\bm{V}} \llbracket f_V(\mathrm{pa}_V) - v \rrbracket  \right]\,,
\\ \label{eq:ll1}
LL_{S=1}(\theta_{\bm{U}})&:=&\sum_{\bm{v} \in \mathcal{D}_1} \log 
\left[\prod_{U\in\bm{U}} \theta_{u} \cdot 
\prod_{V\in\bm{V}} \llbracket f(\mathrm{pa}_V) - v \rrbracket  \right]
\,.
\end{eqnarray}

The following result provides a generalisation of Proposition~\ref{prop:iff} to the case of biased data.

\begin{theorem}\label{th:unimodal}
The log-likelihood in Equation~\eqref{eq:llu} has no local maxima and achieving its global maximum at $\theta_{\bm{U}}$ is an equivalent condition for $\theta_{\bm{U}}\in\Theta_{\bm{U}}$.
\begin{proof}
Let us first define the following function:
\begin{equation}\label{eq:h}
h(\bm{v}) := \left\{ \begin{array}{ll} \bm{v}&\mathrm{if}\, g(\bm{v})=1\,,\\ *& \mathrm{otherwise}\,.\end{array}\right.
\end{equation}
Variable $T:=h(\bm{V})$ takes its values from $\Omega_{T}:= \Omega_{\bm{V}}^{S=1} \cup \{*\}$ where:
\begin{equation}
\Omega_{\bm{V}}^{S=1}:=\left\{ \bm{v} \in \Omega_{\bm{V}} : g(\bm{v}) = 1\right\}\,.
\end{equation}
Let $f:=\Omega_{\bm{U}}\to \Omega_{\bm{V}}$ denote the joint representation of the SEs $\{f_V\}_{V\in\bm{V}}$. From $f$ and $h$ we can build a map $l:\Omega_{\bm{U}}\to\Omega_{T}$ by composition, i.e.,$l := h \circ f$. This allows to obtain from $M$ a PSCM $M'$ with the same exogenous variables of $M$, and $T$ being the only endogenous variable with $l$ as SE. Note that $l$ satisfies surjectivity and $M'$ is therefore a proper PSCM such that $T$ is a common child of all the nodes associated with the variables in $\bm{U}$. We similarly obtain from $\mathcal{D}_S$ a dataset $\mathcal{D}'$ of $N$ complete observations of $T$ as follows: for each (missing) observation of $\bm{V}$ in $\mathcal{D}_{S=0}$ (whose number is $N_{S=0}$) we add an observation $T=*$ to $\mathcal{D}'$ (remember that $*$ is a proper state of $T$); while the observations of $\bm{V}$ in $\mathcal{D}_{S=1}$ are directly added to $\mathcal{D}'$ and regarded as observations of $T$ instead of $\bm{V}$. 

The log-likelihood of $\mathcal{D}'$ from $M'$ can be computed as in Equation~\eqref{eq:liku}, i.e.,
\begin{equation}\label{eq:llprime}
LL'(\theta_{\bm{U}}):=\sum_{\bm{t}\in\mathcal{D}'} \log \sum_{\bm{u}\in\Omega_{\bm{U}}} \left[ \prod_{U\in\bm{U}} \theta_u \cdot \llbracket l(\bm{u})-t \rrbracket \right]\,.
\end{equation}
Let us show that Equation~\eqref{eq:llprime} coincides with Equation~\eqref{eq:llu}. Because of Equation~\eqref{eq:h}, the elements of $\mathcal{D}'$ such that $T=*$ are those corresponding to $S=0$, their number being therefore $N_{S=0}$. Their contribution to the log-likelihood is therefore the term in Equation~\eqref{eq:ll0}. For the other elements of $\mathcal{D}'$, function $h$ in Equation~\eqref{eq:h} acts as the identical map and hence $l(\bm{u})=f(\bm{u})$. The corresponding contribution to the log-likelihood is the one in Equation~\eqref{eq:ll1}. This proves that the log-likelihood of a biased dataset in Equation~\eqref{eq:liku} can be expressed as the log-likelihood of an unbiased dataset (namely $\mathcal{D}'$) from a proper PSCM (that is $M'$). This allows to apply Proposition~\ref{prop:iff} and hence to prove the thesis.
\end{proof}
\end{theorem}

As a consequence of Theorem~\ref{th:unimodal}, we can use Algorithm~\ref{alg:cem}
to PSCM $M^S$ and dataset $\mathcal{D}_S$ to obtain FSCMs whose log-likelihood takes its global maximum and hence satisfies the compatibility constraints. Causal queries computed with those FSCMs are consequently inner points of the exact interval for the partially identifiable query. The more points the better the approximation. 

Note that as the data in $\mathcal{D}_{S=0}$ are $N_{S=0}$ instances of the same observation $S=0$. Thus, when coping with biased data, line 5 of Algorithm~\ref{alg:cem} rewrites as:
\begin{equation}\label{eq:newline}
P_{t+1}(U) \leftarrow \dfrac{N_{S=0} P_t(U|S=0)+\sum_{\bm{v} \in \mathcal{D}_1} P_t(U|\bm{v})}{
N_{S=0}+N_{S=1}}\,.
\end{equation}

Equation~\eqref{eq:newline} shows the divergent effect of two datasets $\mathcal{D}_{S=0}$ and $\mathcal{D}_{S=1}$ on the exogenous chances. While the selected states push the chances towards their unbiased values, the unselected ones act as a noise term only forcing the chances to be zero on the exogenous states generating the selected endogenous states. Overall, compared to the unbiased case, we expect this weighted average to induce larger bounds on the counterfactual queries. An example is reported here below.

\begin{figure}[htp!]
\centering
\pgfplotstableread{exe1_qm.dat}{\tablea}
\pgfplotstableread{exe2_qm.dat}{\tableb}
\begin{tikzpicture}[]
\pgfplotsset{every x tick label/.append style={font=\footnotesize, yshift=-0.5ex}}
\pgfplotsset{every y tick label/.append style={font=\footnotesize, yshift=-0.5ex}}
\begin{axis}[xmin = 0, xmax = 1, ymin = 0, ymax = 1,
xlabel = {$p(S=1)$},ylabel = {PNS},
x tick label style={rotate=330,anchor=west},
xtick = {1.,0.895,0.6395,0.5,0.15,0.0},
ytick = {0.250,0.500,0.750,1.000},
yticklabels = {0.25,0.50,0.75,1.00},
xticklabels = {1.00,0.89,0.64,0.50,0.15,0.00},
grid = both,major grid style = {lightgray},minor grid style = {lightgray!25}]
\addplot[gray,fill,opacity=0.2,sharp plot] table [x ={x}, y = {upper}] {\tablea}|- (current plot begin);
\addplot[black, only marks, mark size=1pt] table [x ={x1}, y = {y1}]{\tableb};
\addplot[black, only marks, mark size=1pt] table [x ={x2}, y = {y2}]{\tableb};
\addplot[black, only marks, mark size=1pt] table [x ={x3}, y = {y3}]{\tableb};
\addplot[black, only marks, mark size=1pt] table [x ={x4}, y = {y4}]{\tableb};
\addplot[black, only marks, mark size=1pt] table [x ={x5}, y = {y5}]{\tableb};
\addplot[black, only marks, mark size=1pt] table [x ={x6}, y = {y6}]{\tableb};
\end{axis}
\end{tikzpicture}
\caption{PNS bounds for different selection levels (x-axis). Bounds (grey) induced by $r=300$ EMCC runs and the points associated with each run (black) are depicted.}
\label{fig:sb2}
\end{figure}
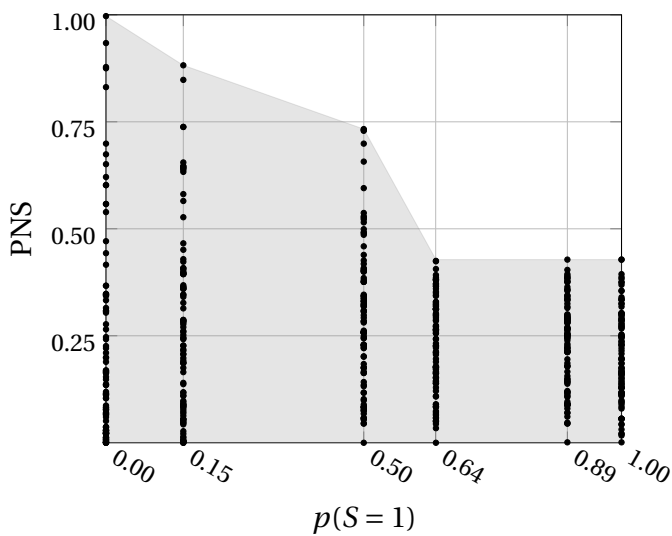

\begin{example}\label{ex:biased}
Consider the same setup as in Example~\ref{ex:unbiased}. Here we compute the PNS for the observational data in Table~\ref{tab:study} with a selector preventing the data associated with the grey rows from being available. These are half of the data, which means $P(S=1)=0.5$. With the two datasets induced by such a selector (see Figure~\ref{fig:sb}), the modified version of Algorithm~\ref{alg:cem} for biased data returns the interval $[0.00,0.73]$. Figure~\ref{fig:sb2} depicts the PNS intervals (and the points spanning these intervals) for different selection mechanisms corresponding to different values of $P(S=1)$. The size of the PNS interval for the unbiased case increases for decreasing values of $P(S=1)$, this reflecting the fact that we focus on selection mechanisms obtained by removing records incrementally.
\end{example}

Note that in the limit $P(S=1)\to 0$, we are basically ignoring the data in $\mathcal{D}_{S=1}$. If this is the case, the log-likelihood is proportional to $N_{S=0}$ and this implies that EMCC gives the same results irrespectively of the size of the (completely missing) dataset. In those cases our EMCC trivially returns the exogenous PMFs sampled for initialisation. This explains, for instance, the (almost) vacuous bounds obtained in this limit in the left side of Figure~\ref{fig:sb2}.

Let us also remark, with reference to Equation~\eqref{eq:newline}, that the execution of EMCC under selection bias requires the cardinality $N_{S=0}$ of $\mathcal{D}_{S=0}$. Yet, as the data in $\mathcal{D}_0$ are unavailable by definition, $N_{S=0}$ might be unavailable too. If this is the case, it is in principle possible to estimate $N_{S=0}$ from $P(S=0)$ as $N_{S=0}/N_{S=1} \simeq P(S=0)/P(S=1)$. If also $P(S=0)$ is not available, a conservative approach may consist in using an upper bound for $P(S=0)$. If this is not possible either, one can still use the (very) conservative approach of taking the limit $P(S=0)\to 1$.

Finally notice that as in principle $S$ might be a common child of all the variables in $\bm{V}$, implementing SE $g$ as a CPT might lead to an exponential blow-up, this making the inference of $P(U|S=0)$ required by EMCC intractable. In practice, as in \citet{bareinboim2012controlling} or in our example, the selector might only depend on a subset of $\bm{V}$ of bounded cardinality. Circuital approach like the one recently proposed by \citet{darwiche2022causal} might be considered to bypass such a limit.

\section{Coping with Multiple Datasets}\label{sec:multi-db}
In the previous sections we addressed the problem of partial identifiability in causal queries when coping with a purely observational dataset, no matter whether unbiased (Section~\ref{sec:emcc}) or biased (Section~\ref{sec:s-emcc}). The challenge of this section is to extend our approach to the case of multiple datasets in a fully general setup mixing observational and interventional data. We initially assume the data unbiased just for the sake of presentation. The extension to the biased case is discussed in the last part of the section.

Say that our domain of interest is described by a PSCM $M$ over $(\bm{U},\bm{V})$. Endogenous observations from $d$ \emph{independent}, observational or interventional, studies are available. Denote as $\mathcal{D}^k$ the dataset of size $N^k:=|\mathcal{D}^k|$ with the results of the $k$-th study, for each $k\in\{1,\ldots,d\}$. The endogenous variables $\bm{V}_I^k\subseteq\bm{V}$ are those subject to interventions in $\mathcal{D}^k$, while for observational studies we simply set $\bm{V}_I^k:=\emptyset$. Without any lack of generality, we assume that the intervened variables are always the same across the records of a same dataset. If this is not the case, we just split the dataset into smaller groups homogeneous w.r.t. the set of intervened variables. The states induced by the interventions inside each dataset might instead be different (e.g., a clinical study where we intervene by giving or {not} giving the drug to a patient). For $\mathcal{D}^k$, we denote these states as $\tilde{\Omega}_{\bm{V}_I^k} \subseteq \Omega_{\bm{V}_I^k}$, while $\tilde{\Omega}_{\bm{V}_I^k}:=\emptyset$ for an observational $\mathcal{D}^k$. We further clarify such a notation by means of the following example.

\begin{example}\label{ex:setup}
Let $\mathcal{D}^1$ denote the interventional dataset in Table~\ref{tab:study} and $\mathcal{D}^2$ the (unbiased) observational one. We have $\bm{V}^1_I=\{ \mathrm{Treatment} \}$, $\tilde{\Omega}_{\bm{V}_I^1}=\{ \mathrm{drug}, \, \mathrm{no\,drug}\}$, $\bm{V}^2_I=\emptyset$ and $\tilde{\Omega}_{\bm{V}_I^2}=\emptyset$.
\end{example}

We merge the $d$ datasets into a single one, say $\mathcal{D}:=\cup_{k=1}^d \mathcal{D}^k$, with cardinality $N:=\sum_{k=1}^d N^k$. We introduce a new variable $W$ to index all the interventional states  in $\mathcal{D}$. An additional state of $W$, denoted as $w_{\emptyset}$, describes the case of no intervention. Overall we have $\Omega_W := \cup_{k=1}^d \tilde{\Omega}_{\bm{V}_I^k} \cup \{ w_{\emptyset} \}$. We consequently augment $\mathcal{D}$ with a column of observations of $W$. The values of $W$ for records originally belonging to the interventional datasets are those corresponding to the intervened states, while the state $w_{\emptyset}$ is assigned to the records from the observational datasets. We eventually denote as $\mathcal{D}'$ the dataset of $N$ (complete) observations of $(\bm{V},W)$ obtained in this way. Let us demonstrate the above merging procedure by means of an example.

\begin{example}\label{ex:merged}
Consider the setup in Example~\ref{ex:setup} with $\mathcal{D}:=\mathcal{D}^1\cup\mathcal{D}^2$. The index variable $W$ is such that $\Omega_W:=\{ \mathrm{drug}, \mathrm{no}\,\mathrm{drug}, w_{\emptyset} \}$. The corresponding augmented dataset $\mathcal{D}'$ is in Table~\ref{tab:study-augmented}.
\end{example}

\begin{table}[htp!]
\centering
\begin{tabular}{llllr}
\toprule
Treatment&Gender&Survival&$W$&Counts\\
\midrule
drug&female&survived&drug&489\\
drug&female&dead&drug&511\\
drug&male&survived&drug&490\\
drug&male&dead&drug&510\\
no drug&female&survived&no drug&210\\
no drug&female&dead&no drug&790\\
no drug&male&survived&no drug&210\\
no drug&male&dead&no drug&790\\
drug&female&survived&$w_{\emptyset}$&378\\
drug&female&dead&$w_{\emptyset}$&1022\\
drug&male&survived&$w_{\emptyset}$&980\\
drug&male&dead&$w_{\emptyset}$&420\\
no drug&female&survived&$w_{\emptyset}$&420\\
no drug&female&dead&$w_{\emptyset}$&180\\
no drug&male&survived&$w_{\emptyset}$&420\\
no drug&male&dead&$w_{\emptyset}$&180\\
\bottomrule
\end{tabular}
\caption{A merged version of the two datasets in Table~\ref{tab:study} with the index variable $W$.}
\label{tab:study-augmented}
\end{table}

\begin{figure}[htp!]
\centering
\begin{tikzpicture}[scale=1.0]
\node[dot2,label=above left:{$U_1$}] (u1)  at (2.5,1) {};
\node[dot2,label=above left:{$U_2$}] (u2)  at (-1.5,1) {};
\node[dot,label=above:{$V_1$}] (v1)  at (1,1) {};
\node[dot,label=below left:{$V_2$}] (v2)  at (-1,-1) {};
\node[dot,label=below right:{$V_3$}] (v3)  at (3.0,-1) {};
\node[dot,label=above right:{$W$}] (w)  at (-0.5,2.5) {};
\node[dot2,label=above left:{$U_W$}] (u4)  at (-1,3.5) {};
\draw [a2] (w) .. controls (3,2) .. (v3);
\draw[a2] (w)  -- (v1); 
\draw[a2] (w) -- (v2);
\draw[a] (u2) -- (v2);
\draw[a] (u4) -- (w);
\draw[a] (u1) -- (v1);
\draw[a] (u1) -- (v3);
\draw[a2] (v2) -- (v1);
\draw[a2] (v1) -- (v3);
\draw[a2] (v2) -- (v3);
\end{tikzpicture}
\caption{The auxiliary model $M'$ obtained from the PSCM $M$ in Example~\ref{ex:unbiased} and the merged dataset in Table~\ref{tab:study-augmented}.}\label{fig:scm-augmented}
\end{figure}
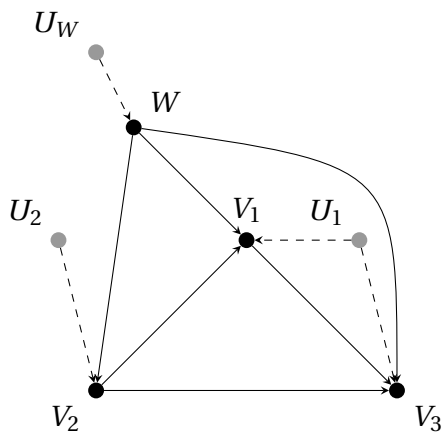

Given $M$ and $\mathcal{D}'$, let us build a, so-called \emph{auxiliary}, PSCM $M'$ as follows. Add variable $W$ to $M$ as an endogenous auxiliary variable corresponding to an additional parent of all the endogenous variables in $\bm{V}$. The SE $f_V'$ associated with $V$ in $M'$ is defined as:
\begin{equation}\label{eq:auxi}
f_V'(\mathrm{pa}_V,W=w)
:=
\left\{
\begin{array}{cc}
v_w & \mathrm{if}\quad V \in \bm{V}_I^{k(w)}\,,\\ 
f_V(\mathrm{pa}_V) & \mathrm{otherwise}\,,\\
\end{array}
\right.
\end{equation}
for each $\mathrm{pa}_V \in \Omega_{\mathrm{Pa}_V}$, $w\in\Omega_W$, and $V\in\bm{V}$, where $f_V$ is the SE of $V$ in $M$, and, $k(w)$ is the index of the interventional dataset associated with $w$, for each $w \neq w_{\emptyset}$, and $v_w\in\Omega_V$ is the state of $V$ appearing in $w$. In practice, Equation~\eqref{eq:auxi} implements the surgery on $V$ required by an intervention if $W$ takes a value corresponding to an intervention involving also $V$, while leaving the original SE of $M$ otherwise. For a proper PSCM specification, we finally add to $M'$ an exogenous variable $U_W$ defined as a unique parent of $W$, with $\Omega_{U_W}:=\Omega_W$ and the identical map as SE for $W$. As an example, Figure~\ref{fig:scm-augmented} depicts the graph of the auxiliary PSCM $M'$ obtained from the PSCM $M$ discussed in Example~\ref{ex:unbiased} and the dataset $\mathcal{D}'$ in Table~\ref{tab:study-augmented}. 

The auxiliary PSCM $M'$ embeds all the possible interventions included in the $d$ datasets. Thanks to the index variable $W$, which takes track of the different interventions, we can regard $\mathcal{D}'$ as a purely observational dataset for the endogenous variables of $M'$. By Proposition~\ref{prop:iff}, we can therefore apply the EMCC scheme to address non-identifiable queries as in Section~\ref{sec:emcc} even when coping with the generalised setup discussed in this section.

In particular, the global maximum of the log-likelihood achieved by the EM scheme should correspond to the log-likelihood assigned to the dataset by an endogenous BN whose parameters have been trained from the dataset itself. As $W$ is a common parent of all the endogenous variables, the endogenous graph of $M'$ is just the endogenous graph of $M$ augmented by $W$, which acts as a common parent of all the other endogenous variables. As a first example of a partially identifiable query based on heterogeneous data let us consider the following example, where the data integration induce tighter bounds.

\begin{example}\label{ex:hybrid}
The PNS query for the (unbiased) observational data in Table~\ref{tab:study} is $[0.00,0.43]$. By also considering the interventional data, the EMCC applied to the model in Figure~\ref{fig:scm-augmented} and the dataset in Table~\ref{tab:study-augmented} returns instead $[0.32,0.42]$. 
\end{example}

So far we only considered the case of unbiased datasets. In the presence of a selection bias, even if only on some datasets, we should first add the selector variable $S$ to all the $d$ datasets, with its value being constantly true on the unbiased ones. The construction of the auxiliary PSCM $M'$ from $M$ is analogous, provided that the auxiliary variable $W$ becomes also a parent of $S$, with the SE of $S$ in $M'$ implementing the different selection functions depending on the particular value of $W$. This requires $S$ to be a common child of all the endogenous variables. A more compact approach consists in considering only the union of the endogenous variables involved in the different biases for the different datasets. Other approaches involving auxiliary variables might be also considered. Note also that we might need different states of $W$ to model the same intervention (or lack of intervention) in different datasets. We do not explicitly formalise this point just for the sake of light notation. Once this is done, Algorithm~\ref{alg:cem} can be executed on $M'$ with $\mathcal{D}'$ as in the unbiased case. An example is reported here below.

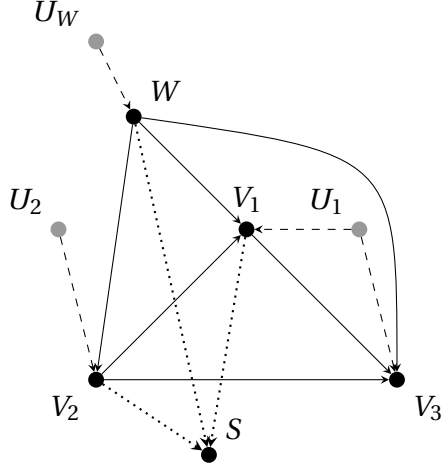
\begin{figure}[htp!]
\centering
\begin{tikzpicture}[scale=1.0]
\node[dot2,label=above left:{$U_1$}] (u1)  at (2.5,1) {};
\node[dot2,label=above left:{$U_2$}] (u2)  at (-1.5,1) {};
\node[dot,label=above:{$V_1$}] (v1)  at (1,1) {};
\node[dot,label=below left:{$V_2$}] (v2)  at (-1,-1) {};
\node[dot,label=below right:{$V_3$}] (v3)  at (3.0,-1) {};
\node[dot,label=above right:{$W$}] (w)  at (-0.5,2.5) {};
\node[dot2,label=above left:{$U_W$}] (u4)  at (-1,3.5) {};
\draw [a2] (w) .. controls (3,2) .. (v3);
\draw[a2] (w)  -- (v1); 
\draw[a2] (w) -- (v2);
\draw[a] (u2) -- (v2);
\draw[a] (u4) -- (w);
\draw[a] (u1) -- (v1);
\draw[a] (u1) -- (v3);
\draw[a2] (v2) -- (v1);
\draw[a2] (v1) -- (v3);
\draw[a2] (v2) -- (v3);
\node[dot,label=above right:{$S$}] (s)  at (0.5,-2) {};
\draw[a2,thick,dotted] (w) -- (s);
\draw[a2,thick,dotted] (v1) -- (s);
\draw[a2,thick,dotted] (v2) -- (s);
\end{tikzpicture}
\caption{The auxiliary model $M'$ obtained from the PSCM $M$ in Example~\ref{ex:unbiased} and the merged dataset in Table~\ref{tab:study-augmented} in the presence of a selection bias.}\label{fig:scm-augmented2}
\end{figure}

\begin{example}\label{ex:hybrid_biased}
In the same setup of Example~\ref{ex:hybrid} consider also the selection bias preventing the grey rows of the observational study in Table~\ref{tab:study} from being available. To compute the PNS in this case we add the Boolean selector $S$ to all the records and obtain a merged dataset $\mathcal{D}'$ as in Table~\ref{tab:study-augmented} with an additional column associated with $S$, which is always one apart from the four grey rows where it is zero. Figure~\ref{fig:scm-augmented2} depicts the corresponding auxiliary model $M'$. Given $\mathcal{D}'$ and $M'$, the EMCC returns a PNS interval equal to $[0.27,0.53]$ (with the observational data only the interval was $[0.00,0.73]$).
\end{example}

As a final remark, notice that so far we implicitly assumed the different datasets to be generated by the same exogenous chances. Yet, this might not always be the case: consider for instance two clinical studies in different countries where the body mass index has very different distributions over the population. Different chances for the different datasets can be simply introduced to cope with such an extended setup. To model this in the auxiliary PSCM $M"$, we should set the index variable $W$ to be also a parent (or the parent of an auxiliary parent) of the exogenous variables having different chances. Although this is not respecting the standard definition of PSCMs (cf. Section~\ref{sec:background}), $W$ is always observed in $\mathcal{D}'$ and therefore acts as a selector for the exogenous PMF to be updated by the EMCC. Such an index should be also specified in the query of interest. The above setup is demonstrated by means of the following example.

\begin{example}
In the same setup of Example~\ref{ex:hybrid_biased} assume that the two studies in Table~\ref{tab:study} refer to two different exogenous chances associated with the endogenous variable $V_2$ (i.e., \emph{Gender}). Let us modify the auxiliary PSCM $M'$ in Figure~\ref{fig:scm-augmented2} in order to take into account this difference. An arc directly connecting $W$ and $U_2$ would not work as the (three) states of $W$ distinguish between the two interventional states in $\mathcal{D}^1$, which in our assumptions refer to the same exogenous chance. This can be solved by a {coarsening} variable $W':=i(W)$, specified as a child of $W$ and returning the index of the dataset we refer to, i.e., $i(W=w_{\emptyset})=1$ and $i(W=\mathrm{drug})=i(W=\mathrm{no\,drug})=2$. Variable $W'$ is eventually specified as a parent of $U_2$ with  $P(U_2|W'=1)$ and $P(U_2|W'=2)$ modelling the expectations of the exogenous chances in the two datasets. The PNS interval we obtain in this case is $[0.20,0.54]$ for the chances associated with the observational dataset, this corresponding to looser bounds than those of the interval $[0.27,0.53]$ obtained in Example~\ref{ex:hybrid_biased} by forcing the chances to be equal. 
\end{example}

\section{Empirical Validation}\label{sec:experiments}
To evaluate the potential of our approach to the computation of partially identifiable queries with heterogeneous data, we report and discuss the results of extensive tests based on a benchmark of synthetic data and causal models (Section~\ref{sec:synthetic}), and a real-world application of counterfactual analysis with biased data (Section~\ref{sec:triangolo}).

Algorithm \ref{alg:cem} was already implemented within the CREDICI library \citep{credici}, a Java tool for causal inference.\footnote{\href{https://www.github.com/idsia/credici}{\tt github.com/idsia/credici}.} We enhance the library with an implementation of the procedures presented in Sections~\ref{sec:s-emcc} and ~\ref{sec:multi-db}, thus allowing to compute counterfactuals with heterogeneous data. To the best of our knowledge this is the first tool for causal inference in such a general setting. The code to reproduce the experiments is available in a dedicated repository.\footnote{\href{https://www.github.com/IDSIA-papers/2023-IJAR-bias-hybrid}{\tt github.com/IDSIA-papers/2023-IJAR-bias-hybrid}.}

\subsection{Synthetic Data}\label{sec:synthetic}
We use the Erd\"os-R\'enyi scheme to uniformly sample directed acyclic graphs. Parentless nodes are regarded as exogenous variables and the other ones as Boolean endogenous ones. SEs and exogenous cardinalities are obtained by sampling (without replacement) the deterministic relations between each endogenous variable and its endogenous parents, letting the states of the exogenous parents index the relations with $|\Omega_U|\leq 64$ for each $U\in\bm{U}$. From such a PSCM $M$ we sample the \emph{ground-truth} chances, thus obtaining FSCM $M^*$. Overall we generate $220$ models with $|\bm{V}|$ ranging from $5$ to $17$.

For each model we select three endogenous variables to be called, respectively, input, target, and covariate. Following a topological order, we set as input the first variable with an exogenous confounder. The target is a leaf such that there is a directed (endogenous) path connecting the input to the target, and the covariate is a variable belonging to that path. We sample from $M^*$ three datasets of endogenous observations, namely: (i) an observational dataset $\mathcal{D}_O$; (ii) an interventional dataset $\mathcal{D}_I$, with interventions on the input and an equal number of positive and negative values for the intervened variable; (iii) another interventional dataset $\mathcal{D}_{IB}$ with interventions on the covariate and a selector based on the values of input, covariate and target. To avoid very strong or very weak biases, the selectors are designed in order to have $0.25 \leq P(S=1) \leq 0.75$. For each model, the three datasets have the same size, with values ranging from $1'000$ to $5'500$ over the different models in order to ensure $M$-compatibility. Note that for the biased dataset, the size is intended to be the one before the selection process.

We want to evaluate the PNS for the input and the target from the PSCM $M$. This is done by starting from $\mathcal{D}_{IB}$, then adding $\mathcal{D}_O$, and finally also $\mathcal{D}_I$. The EMCC is always executed with $r=300$ runs. 

Before commenting the results, we use the credibility intervals derived by \citet{zaffalon2022}
to evaluate the quality of the EMCC inner approximation with the selected number of runs. We compute the probability of a relative error (at each extreme of the interval) smaller than $\epsilon=0.1$ in our simulations and obtain a median confidence equal to $0.999$ and a lower quartile equal to $0.921$. Those high values support the accuracy of our inference scheme.

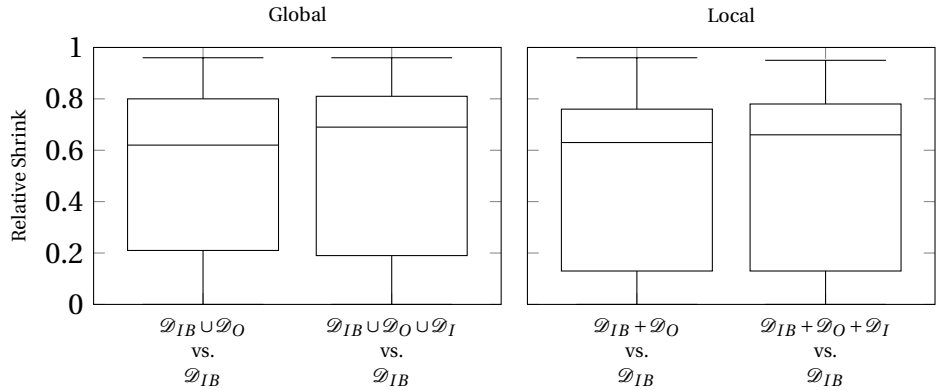
\begin{figure}[ht]
\centering
\begin{tikzpicture}[]
\begin{axis}
[boxplot/draw direction=y,title={\scriptsize Global},ylabel={\scriptsize Relative Shrink},height=5cm,width=7cm,
ymin=0.0,ymax=1.0,cycle list={{black},{black}},
xtick={1,2,3},xticklabels={\scriptsize{$\begin{array}{c}{\mathcal{D}_{IB}\cup\mathcal{D}_{O}}\\\mathrm{vs.}\\{\mathcal{D}_{IB}}\end{array}$},\scriptsize{$\begin{array}{c}{\mathcal{D}_{IB}\cup\mathcal{D}_{O}\cup\mathcal{D}_{I}}\\\mathrm{vs.}\\{\mathcal{D}_{IB}}\end{array}$}}]
\addplot+[fill,fill opacity=0.0,boxplot prepared={median= 0.62 ,upper quartile= 0.80 ,lower quartile= 0.21 ,upper whisker= 0.96 ,lower whisker=0}
] coordinates {};
\addplot+[fill,fill opacity=0.0,boxplot prepared={median= 0.69 ,upper quartile= 0.81 ,lower quartile= 0.19 ,upper whisker= 0.96 ,lower whisker=0.0}] coordinates {};
\end{axis}
\end{tikzpicture}
\begin{tikzpicture}[]
\begin{axis}
[boxplot/draw direction=y,title={\scriptsize Local},ylabel={},height=5cm,width=7cm,
ymin=0.0,ymax=1.0,cycle list={{black},{black}},yticklabels={},
xtick={1,2,3},xticklabels={\scriptsize{$\begin{array}{c}{\mathcal{D}_{IB}+\mathcal{D}_{O}}\\\mathrm{vs.}\\{\mathcal{D}_{IB}}\end{array}$},\scriptsize{$\begin{array}{c}{\mathcal{D}_{IB}+\mathcal{D}_{O}+\mathcal{D}_{I}}\\\mathrm{vs.}\\{\mathcal{D}_{IB}}\end{array}$}}]
\addplot+[fill,fill opacity=0.0,boxplot prepared={median= 0.63 ,upper quartile= 0.76 ,lower quartile= 0.13 ,upper whisker= 0.96 ,lower whisker=0}] coordinates {};
\addplot+[fill,fill opacity=0.0,boxplot prepared={median= 0.66 ,upper quartile= 0.78 ,lower quartile= 0.13 ,upper whisker= 0.95 ,lower whisker=0}
] coordinates {};
\end{axis}
\end{tikzpicture}
\caption{Relative shrink induced by integration of additional datasets.}\label{fig:exp}
\end{figure}

Concerning the results with the heterogeneous data, we notice that adding a new dataset typically induces a shrink in the interval. To summarise the results of our experiments we therefore compute the \emph{relative shrink} defined as one minus the ratio between the size of the PNS interval with the additional dataset(s) and the PNS interval with the initial dataset. Figure~\ref{fig:exp} depicts the corresponding boxplots. On the left we consider the model with the same chances shared over the different datasets (called \emph{global}), while for the \emph{local} models on the right we let the chances associated with the covariate to be different for each dataset, and we run the PNS query with the chances based on $\mathcal{D}_{IB}$.

We observe noticeable shrinks w.r.t. the initial PNS interval, these being above $60\%$ for both global and local models. This advocates the benefits of developing tools for counterfactual inference with heterogeneous data. Such a high shrink is achieved by adding a single dataset to the initial one. The effect of adding the second dataset is considerably weaker (about $5\%$ for both class of models), but still noticeable.

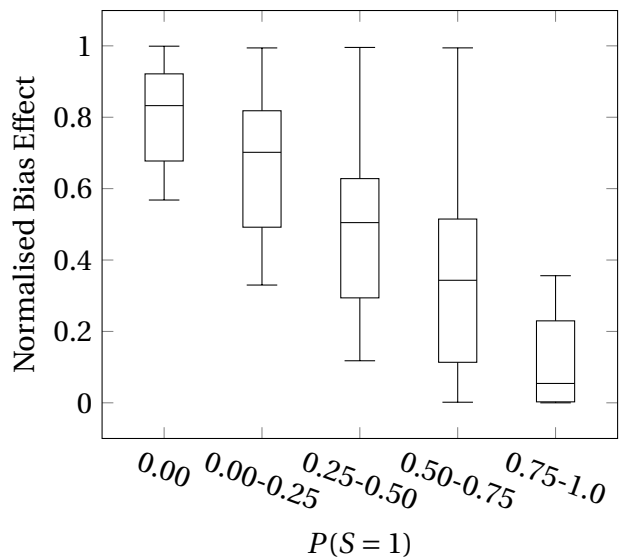
\begin{figure}[htp!]
\centering
\begin{tikzpicture}
\begin{axis}[
boxplot/draw direction=y,
name=border,
ylabel={Normalised Bias Effect},
boxplot={draw position={1/3 + floor(\plotnumofactualtype/3) + 1/3*mod(\plotnumofactualtype,3)},box extend=0.13},
every axis plot/.append style={fill,fill opacity=0.0},
x tick label style={align=center,rotate=-20},
xtick={1/3,2/3,3/3,4/3,5/3,6/3},
xlabel={$P(S=1)$},
xticklabels={0.00,0.00-0.25,0.25-0.50,0.50-0.75,0.75-1.0},
cycle list={{black},{black},{black},{black},{black},{black}}]
\addplot+ table[row sep=\\,y index=0]{data\\0.878352819\\0.964670447\\0.78673313\\0.999037369\\0.568087003\\};
\addplot+ table[row sep=\\,y index=0] {data\\0.749493977\\0.886850415\\0.653996167\\0.994202377\\0.329903232\\};
\addplot+ table[row sep=\\,y index=0] {data\\0.539433512\\0.716697141\\0.470182584\\0.995551487\\0.117807626\\};
\addplot+ table[row sep=\\,y index=0] {data\\0.461162214\\0.568472492\\0.225357124\\0.994310909\\0.001657255\\};
\addplot+ table[row sep=\\,y index=0] {data\\0.10311022\\0.356093195\\0.00540237\\0.812313949\\0\\};
\end{axis}
\end{tikzpicture}
\caption{Difference between the bounds of biased and unbiased data w.r.t. $P(S=1)$.}\label{fig:bias}
\end{figure}

Finally let us also report a dedicated analysis of the effect of the missingness induced by a selection bias with a single observational dataset as in Section~\ref{sec:s-emcc}. We consider different selectors with no restrictions on the values of $P(S=1)$. For each dataset we compare the PNS bounds with the narrower bounds obtained by removing the bias. The quality of the EMCC approximation as based on the credible intervals is analogous to the one observed with the heterogeneous data. As a descriptor we use the difference between the two lower bounds and we normalise it with the difference obtained when the biased interval is $[0,1]$. We call this descriptor \emph{normalised bias effect}. We similarly proceed for the upper bounds. The aggregated results are grouped w.r.t. different ranges of $P(S=1)$ and the corresponding boxplots displayed in Figure~\ref{fig:bias}. As expected with less missing records the bounds become more informative and closer to their unbiased values.

\subsection{A Counterfactual Analysis in Palliative Care with Biased Data}\label{sec:triangolo}
\begin{figure}[htp!]
\centering
\begin{tikzpicture}[scale=1.1]
\node[nodo] (s)  at (0,2.0) {\tiny Symptoms};
\node[nodo] (a)  at (3.5,0.0) {\tiny Age};
\node[nodo2] (ap)  at (-3.5,-1.0) {\tiny Awareness (Patient)};
\node[nodo2] (t)  at (-2,-2.0) {\tiny Triangolo};
\node[nodo] (p)  at (0,-1.0) {\tiny Practitioner};
\node[nodo] (h)  at (+2,-2.0) {\tiny Hospital};
\node[nodo] (k)  at (+4.0,-2.0) {\tiny Karnofsky};
\node[nodo] (sf)  at (+5.5,0.4) {\tiny System (Family)};
\node[nodo2] (af)  at (+6,2.0) {\tiny Awareness (Family)};
\node[nodo] (pf)  at (5.0,-4.0) {\tiny Preference (Family)};
\node[nodo] (pp)  at (-3.0,-4.0) {\tiny Preference (Patient)};
\node[nodo3] (d)  at (1.0,-4.0) {\tiny Death};
\draw[arco] (s) -- (ap);
\draw[arco] (s) -- (af);
\draw[arco] (s) -- (t);
\draw[arco] (s) -- (h);
\draw[arco] (h) -- (p);
\draw[arco] (p) -- (t);
\draw[arco] (k) -- (h);
\draw[arco] (k) -- (sf);
\draw[arco] (af) -- (sf);
\draw[arco] (s) -- (k);
\draw[arco] (s) -- (sf);
\draw[arco] (s) -- (af);
\draw[arco] (sf) -- (pf);
\draw[arco] (ap) -- (pp);
\draw[arco] (pp) -- (d);
\draw[arco] (pf) -- (d);
\draw[arco] (a) -- (k);
\draw[arco] (t) -- (d);
\draw[arco] (h) -- (d);
\end{tikzpicture}
\caption{The graph of the model used to study preferences about the place of death in cancer patients.}
\label{fig:triangolo}
\end{figure}
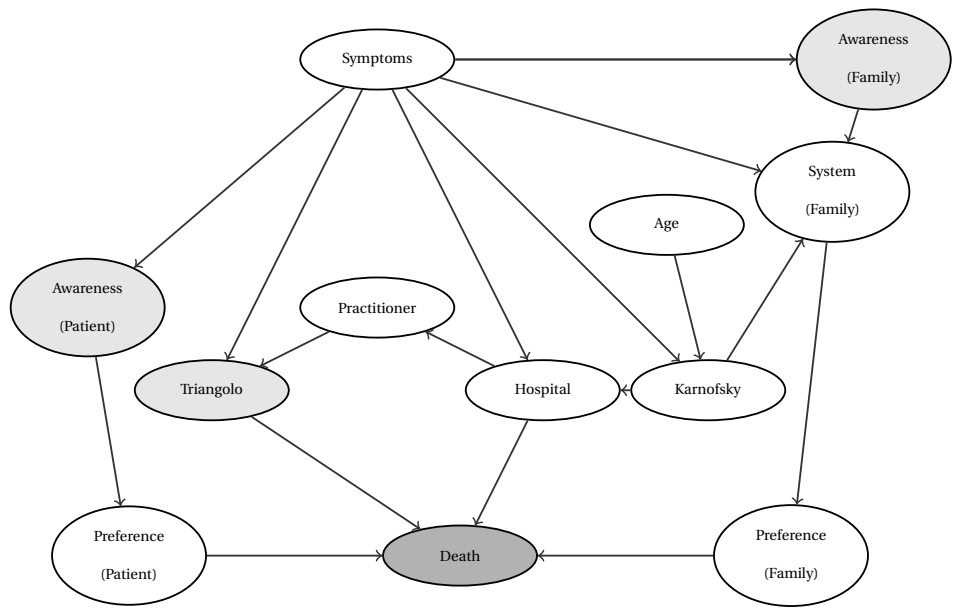

Figure~\ref{fig:triangolo} represents the causal graph used for a study on the preferences of terminally ill cancer patients regarding the place they want to spend their final moments: home or hospital. Although a majority of patients prefer to pass away at home, most of them end up dying in institutional settings. The study focuses on exploring the interventions that healthcare professionals can take to increase the chance of patients dying at home. The graph corresponds to the BN proposed by \cite{kern2020impact} reduced to the subset of variables for which data were available---variables have been binarised too. Due to ethical reasons, the original data of the study cannot be used and we sampled instead a dataset $\mathcal{D}$ of $1'000$ observations. 

One can intervene on three variables in the network (light grey nodes in Figure~\ref{fig:triangolo}): the patient's and the family's \emph{awareness} of death (which involves communication with the doctors); and home assistance (provided by the \emph{Triangolo} association). The model is taken without exogenous confounders as a consequence of the fact that all the potential confounders have been explicitly represented in the causal graph. In practice we consider a \emph{Markovian} model, i.e., each endogenous variable has a separate and unique exogenous parent. As for SEs, we stick to the \emph{canonical} (see \citet{zhang2021}) representation since we want to be least committal w.r.t. the true underlying mechanisms. Note that such a conservative modelling assumption can induce high cardinality for the exogenous variables, in particular the exogenous parent $U$ of variable \emph{Death} has $|\Omega_U|=2^{16}$. Yet, this does not prevent EMCC from running and, in order to evaluate the relative importance of those three variables, we compute the corresponding PNS intervals having those variables as input and \emph{Death} as target node. With the unbiased dataset we obtain:
\begin{eqnarray*}
\mathrm{Triangolo}&\to&[0.27,0.35]\,,\\
\mathrm{Patient\,\,Awareness}&\to&[0.03,0.11]\,,\\
\mathrm{Family\,\,Awareness}&\to&[0.04,0.11]\,. 
\end{eqnarray*}
The first interval clearly dominates the other two, this showing that home assistance is the key factor determining the death place. 

The above result has been presented in a paper under review. Here we investigate whether or not such an evidence in favour of home assistance holds even in the presence of a selection bias. In particular we want to model the fact that studies of this kind are typically biased towards patients progressed to severe conditions. We consequently consider selectors based on \emph{Karnofsky} and \emph{Symptoms}. For both these binary variables we consider the \emph{bad-condition} states, i.e., low Karnofksy index and serious symptoms. We consequently consider a weak bias removing only the records of patients such that both variables are in the bad-condition state, and a strong bias such that even a single variable in a bad-condition state makes the datum unavailable. The results are:
\begin{eqnarray*}
&&\quad\quad\,\mathrm{weak}\,\,\,/\,\,\,\mathrm{strong}\\
\mathrm{Triangolo}&\to&[0.29,0.35] \,/\, [0.22,0.43]\,,\\
\mathrm{Patient\,\,Awareness}&\to&[0.02,0.11] \,/\, [0.02,0.19]\,,\\
\mathrm{Family\,\,Awareness}&\to&[0.04,0.12] \,/\, [0.00,0.19]\,. 
\end{eqnarray*}
In practice even in the presence of a selection bias, no matter whether weak or strong, it is possible to draw informative counterfactuals from the data and keep advocating the importance of home assistance.

Let us finally note that the fact that the lower bound of the PNS for \emph{Triangolo} with a weak bias is slightly higher than the corresponding lower bound in the unbiased case reflects the fact that, with fewer complete data, the (inner) approximation provided by the EMCC with the current number of runs might be stronger than the one obtained with the complete data.

\section{Conclusions and Outlook}\label{sec:conclusions}
We have presented an EMCC algorithm that learns the parameters of a given partially specified structural causal model from a mix of observational and interventional, as well as possibly biased, data. This setting is essentially that of information fusion put forward by \cite{bareinboim2016causal}. In this setting, we appear to be the first to deliver an algorithm with this type of generality and to make the code public. 

A few remarks may be useful to clarify some main aspects of our approach:
\begin{itemize}

\item Our algorithm aims at reconstructing the uncertainty about the latent variables. It does this in an approximate way, by  delivering a set of fully specified SCMs contained in all those that are compatible with the partially specified one. Once this is done, one can compute any counterfactual by iterating its computation over the FSCMs above and aggregating their results so as to yield inner bounds on unidentifiable queries. 

\item Empirical results support the quality of the approximation and confirm the increased informativeness of bounds obtained by merging multiple studies. Note that more efficient, or accurate, computation would be possible in principle if the aim was to compute a specific counterfactual from the beginning, because searching the space of compatible FSCMs could be tailored to such a goal; but this is not the purpose of our work, which aims at remaining agnostic w.r.t. the follow-up computations, and hence general.

\item The PSCM is an input for our algorithm. We stress that the requirement that SEs are given (they are part of the PSCM) is not as stringent as it might seem, given that we can produce the needed SEs by a preprocessing step if the actual ones are not available: this is possible thanks to recent work \citep{duarte2021,zhang2021} that has introduced a \emph{canonical} specification of the SEs. It can be understood as a least-committal specification that, loosely speaking, can be used without loss of generality (the implication however is that the output intervals will tend to be weaker compared to the case where the actual SEs are given). In this sense, our work is therefore as general as the works that do not assume the SEs to be given.

\end{itemize}

As for future work, we would like to compare our EMCC with the MCMC approach recently put forward by \citet{zhang2021}. At the moment it is not clear to us, from the reported experiment, the extent to which their approximate scheme handles heterogeneous data, and with which accuracy, nor whether their method permits checking $M$-compatibility among studies.\footnote{The related code does not appear to be publicly available yet.} This is important because $M$-incompatibility renders inference not tenable.

\section*{Acknowledgements}
\indent The authors are grateful to Heidi Kern from the Triangolo association for her support with the palliative care problem discussed in Section~\ref{sec:triangolo}. This research was partially funded by MCIN/AEI/10.13039/501100011033 with FEDER funds for the project PID2019-106758GB-C32, and also by Junta de Andaluc\'{i}a grant P20-00091. Finally we would like to thank the ``Mar\'{i}a Zambrano'' grant (RR\_C\_2021\_01) from the Spanish Ministry of Universities and funded with Next Generation EU funds.
\bibliographystyle{elsarticle-harv} 
\bibliography{biblio}
\end{document}